\documentclass[journal]{IEEEtran}
\usepackage[T1]{fontenc}
\usepackage{amsmath,amsfonts,amssymb}
\usepackage{array}
\usepackage{graphicx}
\graphicspath{{./}{MPIG/}}
\usepackage[caption=false,font=footnotesize]{subfig}
\usepackage{textcomp}
\usepackage{stfloats}
\usepackage{url}
\usepackage{verbatim}
\usepackage{cite}
\usepackage{booktabs}
\usepackage{multirow}
\usepackage[colorlinks=true, citecolor=blue, linkcolor=blue, urlcolor=blue]{hyperref}
\begin{document}

\title{MPIG: Multi-angle Polarization Image Generation from a Single Polarization Observation}

\author{Jue Wang, Ruilin Huang, Liquan Shen, Zhen Tian, Xia Hua
\thanks{This work was supported by NSFC under Grant Nos.62301305, 62671366, the Guangdong Basic and Applied Basic Research Foundation under Grant 2025A1515012083 and the General Funding of the Open Project of the Key Laboratory of Specialty Fiber Optics and Optical Access Network under Grant SKLSFO2024-05.}
\thanks{Jue Wang, Ruilin Huang, Liquan Shen and Xia Hua are with the School of Communication and Information Engineering, Shanghai University, Shanghai, China (e-mail: wangjue0801@shu.edu.cn; hrl3132215071@shu.edu.cn; jsslq@shu.edu.cn; xiahua@shu.edu.cn).}%
\thanks{Zhen Tian is with the Guangdong Laboratory of Artificial Intelligence and Digital Economy (SZ), Shenzhen, China (e-mail: tianzhen@gml.ac.cn).}
}


\maketitle

\begin{abstract}
Polarization imaging provides informative cues for material analysis, scene understanding, and visual perception. However, acquiring complete polarization observations typically requires specialized hardware and multi-angle measurements, thereby increasing system complexity and limiting practical deployment. In this work, we address the problem of predicting the missing 45$^\circ$, 90$^\circ$, and 135$^\circ$ polarization views from a single 0$^\circ$ observation by leveraging a learned scene prior.
To this end, we propose MPIG (Multi-angle Polarization Image Generation), a Swin Transformer-based framework that couples cross-angle feature interaction with residual decoding to predict angle-dependent intensity variations while preserving shared scene structure. Specifically, measured Stokes differences and polarization-strength-aware supervision guide the recovery of polarization modulation, orientation, and magnitude, while low-rank factorization further reduces the decoding cost.
We also construct HighPol, a large-scale real-world polarization dataset comprising 5,474 groups of high-resolution four-angle polarization images collected from diverse indoor and outdoor scenes. Experimental results demonstrate improved image and polarization accuracy, achieving 29.729\,dB PSNR, 0.9505 SSIM, and 21.514$^\circ$ AoLP MAngE while requiring 85.86\,M parameters and 1.126\,T FLOPs. Furthermore, downstream validation on glass and semantic-segmentation tasks suggests that the generated polarization images can preserve task-relevant information, demonstrating the effectiveness of MPIG and the practical utility of predicted polarization observations for downstream vision applications.
\end{abstract}


\section{Introduction}
\IEEEPARstart{P}{olarization} imaging \cite{rodriguez2024pola4all,gao2013gradient,ngo2015shape} has demonstrated substantial value in tasks such as material analysis~\cite{tominaga2008polarization,lin2022image}, autonomous perception~\cite{sun2019multimodal,nguyen20173d}, scene understanding, and polarization-aware visual perception \cite{miyazaki2017surface,muglikar2023event}. Compared with conventional RGB imaging, polarization imaging can reveal the structural properties and material composition of object surfaces, as well as the interactions between light and matter, thereby offering unique advantages in complex visual environments.

\begin{figure}[t]
	\centering
	\includegraphics[width=\columnwidth]{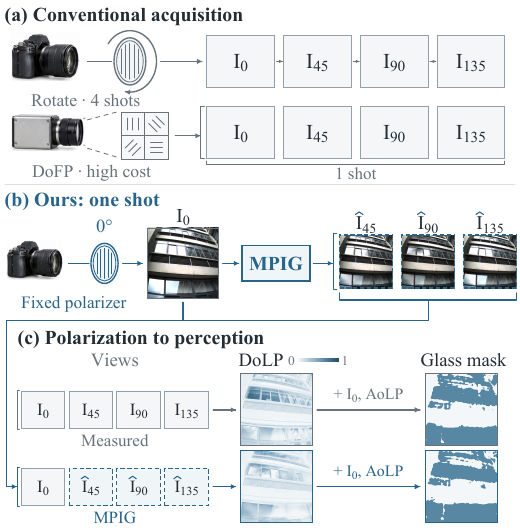}
	\caption{Motivation and practical utility of MPIG.
		(a) Conventional polarization acquisition requires either four
		sequential captures with a rotating polarizer or single-shot
		acquisition using a costly division-of-focal-plane (DoFP) camera.
		(b) The target acquisition setting uses a standard camera with a fixed $0^\circ$
		polarizer to capture one observation $I_0$, from which MPIG
		predicts the missing views $\hat{I}_{45}$, $\hat{I}_{90}$,
		and $\hat{I}_{135}$.
		(c) The completed observations support polarization-assisted
		perception, illustrated by DoLP and glass-segmentation results
		alongside those obtained from four measured views. Both
		segmentation masks are predictions of the same frozen model,
		with blue indicating glass. Acquisition setups and angle
		tiles are schematic.}
	\label{fig:motivation}
\end{figure}

However, the practical adoption of polarization imaging remains restricted by specialized hardware and multi-angle acquisition. As shown in Fig.~\ref{fig:motivation}, existing approaches mainly include sequential rotating-polarizer imaging \cite{huang2000optimizing} and micro-polarizer array cameras \cite{gimenez2020calibration}. The former is difficult to deploy in dynamic or real-time scenarios, whereas the latter increases system complexity and cost. To mitigate these limitations, current research has primarily focused on exploiting incomplete polarization observations for downstream visual tasks, including reflection separation~\cite{yao2025polarfree}, segmentation~\cite{yu2024transparent,mei2022glass}, and transparent-film removal~\cite{tang2024learning}. Although effective in controlled environments, many of these methods still require measurements at multiple analyzer angles to derive polarization cues, which limits their compatibility with conventional cameras. Consequently, broader deployment depends on reducing acquisition requirements and providing suitable data for learning-based prediction.

Meanwhile, existing public polarization datasets are mostly concentrated on specific application scenarios and remain limited in scale, scene diversity, and coverage of highly polarized samples. Such datasets are essential for training and evaluating polarization-assisted methods across diverse materials and environments. Although recent studies \cite{lin2025rgbpolarization,zhang2025polaranything,luo2026genpolar} have attempted to predict polarization information from RGB images, the synthesized observations still exhibit a gap from ground truth (GT), particularly in real-world scenes with complex polarization responses.

To address these issues, we propose MPIG (Multi-angle Polarization Image Generation), a Swin Transformer-based framework~\cite{liu2021swin} that predicts the three missing analyzer views from a single polarization image. A shared hierarchical encoder feeds three angle-specific decoders that predict intensity residuals relative to the observed view. Cross-angle feature interaction exchanges information among correlated responses, while residual prediction preserves shared scene appearance and focuses learning on angle-dependent variations. Measured Stokes differences and polarization-strength-aware supervision guide the recovery of polarization, and low-rank factorization reduces the decoding cost.

We further construct HighPol, a large-scale real-world polarization dataset containing 5,474 groups of high-resolution four-angle images from diverse indoor and outdoor scenes with varied material reflection characteristics. On HighPol, combined with PolarVis from the Sensing And Imaging Lab (SAIL), MPIG achieves 29.729\,dB PSNR and 0.9505 SSIM, exceeding RGB2Pol-MAE by 0.825\,dB while using 74.0\% fewer parameters and 6.8\% fewer FLOPs. Furthermore, semantic and glass-segmentation experiments demonstrate the utility of the predicted polarization cues.

The main contributions of this work are summarized as follows:

\begin{itemize}
	\item MPIG, an efficient framework for completing four-angle polarization observations from a single $0^\circ$ analyzer image.
	\item Cross-angle feature interaction and residual decoding, coupled with measured Stokes and polarization-strength-aware supervision, for recovering angle-dependent intensity variations.
	\item HighPol, a large-scale real-world dataset of 5,474 high-resolution four-angle image groups spanning diverse indoor and outdoor scenes.
\end{itemize}

The remainder is organized as follows. Section~\ref{sec:related_work} reviews related work, Section~\ref{sec:Method} presents MPIG, and Section~\ref{sec:data} describes the dataset and implementation. Section~\ref{sec:experiments} reports the experiments, and Section~\ref{sec:conclusion} concludes the paper.

\section{Related Work}
\label{sec:related_work}

\subsection{Polarization Imaging Systems and Applications}
\noindent\textbf{Polarization Applications.}
Polarization imaging captures surface and material cues that complement conventional intensity measurements. These cues facilitate a wide range of tasks, including surface normal estimation~\cite{wang2026physics,lei2022shape}, reflection separation~\cite{wen2021polarization}, material recognition~\cite{lin2022image}, and transparent object analysis~\cite{yu2024transparent,shao2023transparent}. Specifically, surface reconstruction methods relate polarization responses to object geometry, whereas transparent-object methods exploit polarization-dependent reflections to distinguish glass from its surroundings~\cite{kalra2020deep}. Such information is particularly valuable when appearance alone provides weak boundaries or ambiguous material cues.

To support these applications, several datasets have been developed. For instance, PolaRGB~\cite{yao2025polarfree} provides paired RGB and polarization images from real scenes, RGBP-Glass~\cite{mei2022glass} focuses on transparent object segmentation, and PDAVIS~\cite{mei2023deep} records polarization information with high temporal resolution using an event camera. Their diverse acquisition formats and annotations reflect the requirements of image restoration, object segmentation, and dynamic-scene analysis, respectively. However, these resources are generally tailored to specific tasks and offer limited coverage in terms of scene types, materials, and polarization strengths. Greater scene diversity is therefore essential for studying polarization across both indoor and outdoor environments.

\noindent\textbf{Polarization Acquisition.}
As illustrated in Fig.~\ref{fig:motivation}, rotating-polarizer systems~\cite{huang2000optimizing} combine a conventional camera with sequential analyzer measurements to estimate Stokes parameters. Although such systems are inexpensive, scene motion between exposures may introduce misalignment. In contrast, division-of-focal-plane (DoFP) cameras~\cite{gimenez2020calibration} integrate micro-polarizers at the pixel level, thereby capturing multiple analyzer orientations in a single exposure. This design improves temporal alignment and supports dynamic scenes, but it relies on dedicated hardware and spatial multiplexing. Since each orientation samples only a subset of the sensor grid, interpolation is required to reconstruct complete analyzer images, and the resulting reduced spatial resolution and pixel crosstalk can degrade polarization estimates. Consequently, the two acquisition strategies embody a trade-off between hardware cost and spatial sampling on the one hand, and temporal consistency on the other.

\subsection{Ill-posed Reconstruction and Image Generation}
\noindent\textbf{Learned Reconstruction and Image Generation.}
Recovering high-dimensional scene information from incomplete observations is inherently ill-posed, as exemplified by HDR~\cite{wu2020hdr}, light-field~\cite{vadathya2017learning}, hyperspectral~\cite{zhang2024neural}, and 3D recovery through scattering media~\cite{luo2026polarization}. To alleviate this ill-posedness, learned priors capture statistical regularities that constrain the missing information. For instance, Wu et al.~\cite{wu2018deep} employ network priors to handle blur and motion, whereas Vadathya et al.~\cite{vadathya2017learning} integrate learned priors with structured reconstruction to recover light fields from coded observations. These approaches explicitly link scene statistics with the underlying image formation process.

Image generation architectures offer complementary means of representing such priors. U-Net~\cite{ronneberger2015u} combines an encoder--decoder hierarchy with multi-scale feature fusion to preserve spatial detail. GANs~\cite{goodfellow2014generative,radford2015unsupervised}, including Pix2Pix~\cite{isola2017image}, adopt adversarial training to learn image translation and promote realistic outputs. Swin Transformer~\cite{liu2021swin} aggregates local context via shifted-window attention and constructs hierarchical scene representations. Diffusion models~\cite{ho2020denoising} generate images through iterative denoising; for example, SR3~\cite{saharia2022image} applies conditional diffusion to super-resolution, and Kim et al.~\cite{kim2024diffusion} combine diffusion features with GAN inversion for multimodal face generation. Unlike direct image regression, adversarial training assesses the resemblance between generated and real images, whereas denoising progressively transforms a noisy representation into an image conditioned on the observation. Across these architectures, perceptual realism and pixel fidelity emphasize different aspects of image quality, while the observation model ultimately determines which scene quantities must be inferred.

\noindent\textbf{Polarization Prediction and Generation.}
Recent methods have extended learned image prediction to polarization estimation from conventional RGB inputs. Lin et al.~\cite{lin2025rgbpolarization} compare restoration and generative architectures for RGB-to-polarization estimation, including an estimator based on a pretrained masked autoencoder (MAE), thereby investigating how appearance representations transfer to polarization prediction. PolarAnything~\cite{zhang2025polaranything} employs image-conditioned latent diffusion to predict DoLP and a periodic AoLP representation. Representing the angular quantity periodically accounts for the equivalence of orientations separated by $180^\circ$, and the predicted polarization quantities are subsequently used to synthesize analyzer views. In a different formulation, GenPolar~\cite{luo2026genpolar} estimates channel-wise linear Stokes components, thereby retaining polarization variation across RGB channels, and further distills a diffusion teacher into a one-step generator to reduce the number of inference steps. These studies differ in their output parameterization and generation process: DoLP/AoLP describe polarization magnitude and orientation, whereas Stokes components express intensity and signed modulation. Both representations provide a basis for synthesizing images at specified analyzer angles and connect the estimated polarization state to observable intensity variations.

\section{Method}
\label{sec:Method}

This section first formulates the polarization prediction problem in Section~\ref{subsec:formulation}, then describes shared encoding and angle-specific decoding in Sections~\ref{subsec:encoding} and~\ref{subsec:decoding}, cross-angle interaction in Section~\ref{subsec:acaim}, and image fidelity and polarization-aware supervision in Section~\ref{subsec:objective}.

\begin{figure*}[!t]
	\centering
	\includegraphics[width=\textwidth]{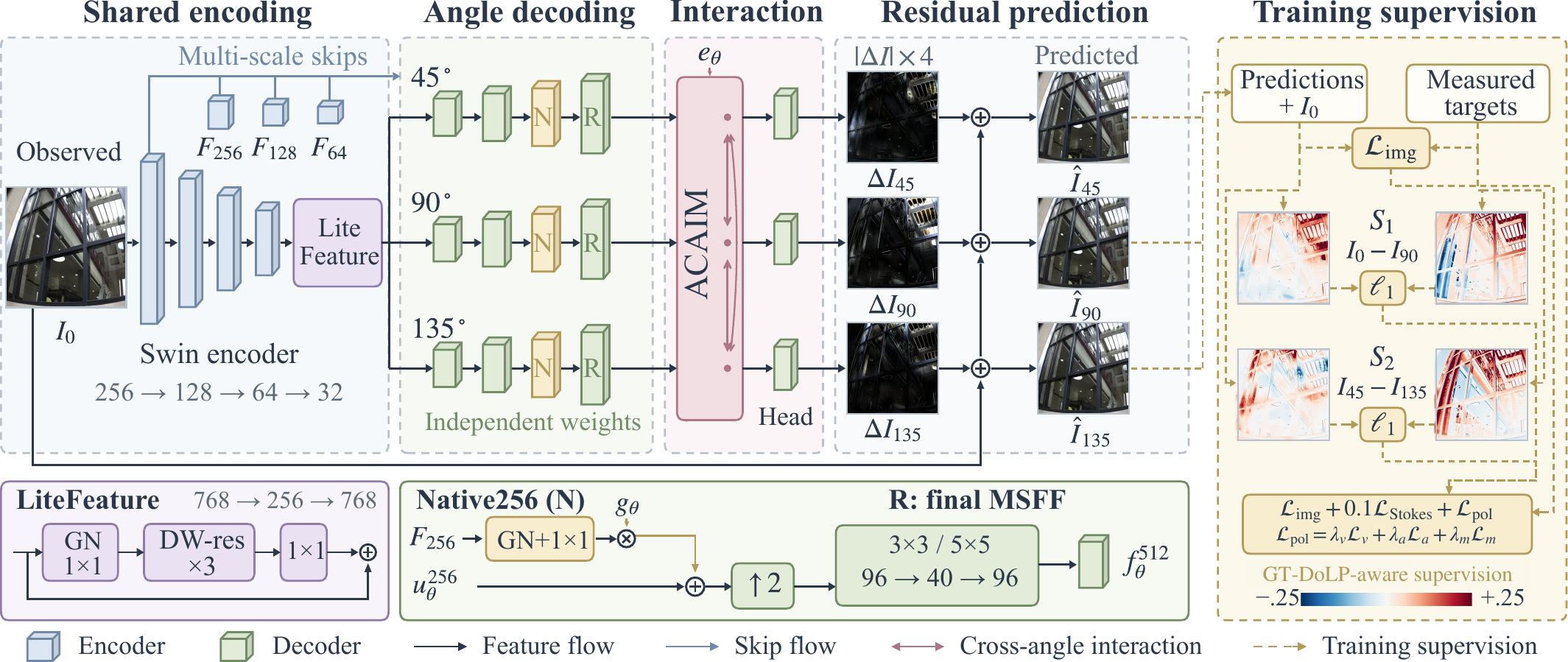}
	\caption{Overview of MPIG. Shared encoding and LiteFeature feed three angle-specific decoders with gated native-resolution skips and rank-40 fusion. ACAIM exchanges cross-angle features before residual prediction. Measured images and Stokes differences supervise image fidelity; normalized-Stokes, direction, and magnitude losses emphasize polarization modulation according to GT polarization strength. Insets show LiteFeature and final-stage decoding.}
	\label{fig:mpig_framework}
\end{figure*}

\subsection{Problem Formulation}
\label{subsec:formulation}

Let $I_\theta\in[0,1]^{3\times H\times W}$ denote an RGB image observed through a linear analyzer at angle $\theta$. Given only $I_0$, we aim to predict $I_\theta$ for $\theta\in\mathcal A=\{45^\circ,90^\circ,135^\circ\}$. Under the ideal linear-polarization model,
\begin{equation}
 I_\theta=\tfrac12(S_0+S_1\cos2\theta+S_2\sin2\theta),
 \label{eq:analyzer}
\end{equation}
where $S_0$ denotes the total intensity, while $S_1,S_2$ characterize the linear polarization. The four analyzer images yield the Stokes parameters as
\begin{align}
 S_0&=I_0+I_{90}, \label{eq:stokes}\\
 S_1&=I_0-I_{90}, \label{eq:stokes_s1}\\
 S_2&=I_{45}-I_{135}. \label{eq:stokes_s2}
\end{align}
For $S_0>0$, the degree and angle of linear polarization are respectively defined as $\rho=\sqrt{S_1^2+S_2^2}/S_0$ and $\psi=\frac12\operatorname{atan2}(S_2,S_1)$, respectively, with $\psi$ taken modulo $\pi$. For an ideal, physically valid polarization state, the observations must additionally satisfy
\begin{align}
 I_0+I_{90}&=I_{45}+I_{135}, \label{eq:ideal_constraints}\\
 S_1^2+S_2^2&\leq S_0^2, \label{eq:stokes_cone}
\end{align}
with $S_0\geq0$. Since $I_0$ constrains $S_0+S_1$ but leaves $S_2$ unobserved, predicting the remaining views is an underdetermined inverse problem. To estimate the missing responses under this ambiguity, MPIG learns appearance–polarization associations from paired observations and exploits spatial context as a data-driven prior.

Accordingly, we parameterize each missing view as
\begin{equation}
 \widehat I_\theta=I_0+\Delta I_\theta,\qquad\theta\in\mathcal A.
 \label{eq:residual_prediction}
\end{equation}
The network predicts residuals in normalized image space and rescales them to obtain $\Delta I_\theta$. Because the analyzer views are spatially registered, this parameterization preserves a common image structure while allocating the learned correction to angle-dependent intensity changes. Under Eq.~\eqref{eq:analyzer}, the ideal residuals satisfy
\begin{align}
 \Delta I_{90}&=-S_1, \label{eq:residual_stokes}\\
 \Delta I_{45}&=\tfrac12(S_2-S_1), \label{eq:residual_stokes45}\\
 \Delta I_{135}&=-\tfrac12(S_1+S_2). \label{eq:residual_stokes135}
\end{align}
The shared dependence of these residuals on $S_1,S_2$ motivates the cross-angle feature interaction in our framework.

\subsection{Shared Encoding and Lightweight Refinement}
\label{subsec:encoding}

As shown in Fig.~\ref{fig:mpig_framework}, a shared Swin-style hierarchical encoder~\cite{liu2021swin} extracts multi-scale features from $I_0$ via patch embedding, window attention, and patch merging. At an input resolution of $512\times512$, it produces features $F_{256}$, $F_{128}$, $F_{64}$, and a latent feature $F_{32}$, where the subscripts denote spatial resolution. Shifted-window attention and patch merging together yield a shared appearance representation, while the high-resolution features retain fine detail for all output angles.

Subsequently, LiteFeature refines $F_{32}$ through a residual bottleneck. A pointwise projection first reduces its width, three depthwise residual blocks then aggregate local context, and a second pointwise projection restores the original width. Denoting SiLU by $\phi$ and the stack of depthwise residual blocks by $\mathcal B$, the refinement is formulated as
\begin{align}
 Z&=\phi(P_\downarrow(\operatorname{GN}(F_{32}))), \label{eq:lite_projection}\\
 \widetilde F_{32}&=F_{32}+P_\uparrow(\mathcal B(Z)),
 \label{eq:litefeature}
\end{align}
where $P_\downarrow$ and $P_\uparrow$ are channel-reducing and channel-restoring pointwise projections, respectively. The zero-initialized output projection initially preserves the encoded features.

\subsection{Efficient Angle-Specific Decoding}
\label{subsec:decoding}

Three independently parameterized decoders progressively upsample the refined latent feature through four stages. At the $64\times64$ and $128\times128$ stages,  the corresponding encoder features are concatenated with the decoder features and fused by pointwise convolutions. This branch-wise independence allows different analyzer responses to be represented within a common scene context. Within each stage, multi-scale feature fusion (MSFF) combines parallel $1\times1$, $3\times3$, and $5\times5$ convolutions. Their outputs are concatenated, reweighted by channel and spatial attention, and summed across scale groups.

To preserve fine structures, Native256 injects the high-resolution encoder feature before the final upsampling stage:
\begin{equation}
 \overline U_\theta^{256}=U_\theta^{256}
 +g_\theta\odot P_{256}(\operatorname{GN}(F_{256})),
 \label{eq:native_skip}
\end{equation}
where $P_{256}$ is a shared pointwise projection and $g_\theta$ is a learned branch-specific channel vector. The projection is initialized to zero, and the gates are initialized to one. The shared projection supplies fine detail, while the channel gates modulate its contribution to each output angle.

The final MSFF operates at full image resolution, making dense spatial convolutions expensive. Its $3\times3$ and $5\times5$ kernels are factorized into a spatial projection from 96 to 40 channels followed by a pointwise projection back to 96, without an intervening activation. For a stage feature $U_\theta$, the factorized branch is
\begin{equation}
 V_{\theta,k}=P_{\theta,k}\bigl(K_{\theta,k}(U_\theta)\bigr),
 \qquad k\in\{3,5\},
 \label{eq:factorization}
\end{equation}
where $K_{\theta,k}$ and $P_{\theta,k}$ denote the spatial and pointwise convolutions, respectively. For a $C$-channel feature and a $k\times k$ kernel, the factorization reduces the multiply--accumulate count from $HW C^2k^2$ to $HW Cr(k^2+1)$, with $r=40$. This rank-40 design, denoted R40, reduces the cost of the two spatial kernels while retaining the pointwise branch, attention operations, and independent angle decoders.

\subsection{Angle-Aware Cross-Angle Interaction}
\label{subsec:acaim}

Although independent decoding preserves angle-specific capacity, it does not explicitly exchange information among the correlated analyzer responses. To address this, ACAIM connects the three branches after spatial decoding and before image prediction, enabling each branch to condition its residual estimate on the other two. Specifically, each target branch employs a learnable code $e_\theta=[\cos2\theta,\sin2\theta]^{\mathsf T}+b_\theta$, where the two-dimensional offset $b_\theta$ is initialized to zero and optimized jointly with the network. For pixel $p$, the encoding is concatenated with the branch feature to form $z_\theta(p)=[f_\theta(p);e_\theta]$.  Shared pointwise projections then produce
\begin{align}
 q_\theta&=W_qz_\theta, \label{eq:acaim_q}\\
 k_\theta&=W_kz_\theta, \label{eq:acaim_k}\\
 v_\theta&=W_vz_\theta. \label{eq:acaim_v}
\end{align}
Omitting $p$ for readability, the cross-angle attention is computed as
\begin{align}
 a_{\theta\varphi}
 &=\frac{\exp(q_\theta^{\mathsf T}k_\varphi/\sqrt d)}
 {\sum_{\eta\in\mathcal A}\exp(q_\theta^{\mathsf T}k_\eta/\sqrt d)},
 \label{eq:acaim_weights}\\
 \widetilde f_\theta
 &=f_\theta+\gamma\sum_{\varphi\in\mathcal A}a_{\theta\varphi}v_\varphi,
 \label{eq:acaim_update}
\end{align}
where $d$ is the query dimension and the learned scalar $\gamma$ starts from zero. At each pixel, ACAIM computes a feature-dependent $3\times3$ attention matrix to exchange information among the analyzer branches. Independent convolutional heads then predict the residuals in Eq.~\eqref{eq:residual_prediction}. Zero-initialized output layers provide an identity initialization for training.

\subsection{Image Fidelity and Polarization-Aware Supervision}
\label{subsec:objective}

The objective illustrated in Fig.~\ref{fig:mpig_framework} combines image fidelity with supervision of measured polarization differences. Let $J_\theta=\mathcal G(I_\theta)$ denote the grayscale intensity, where $\mathcal G$ applies standard RGB luminance weights, and let $m_{b\theta}$ be the grayscale mean squared error for sample $b$ and angle $\theta$. The image objective is defined as
\begin{equation}
 \mathcal L_{\mathrm{img}}=0.01\mathcal L_{\log}
 +0.1\mathcal L_{\mathrm{rgb}}+0.1\mathcal L_{\mathrm{ssim}}.
 \label{eq:image_objective}
\end{equation}
For a batch of size $B$, its components are
\begin{align}
 \mathcal L_{\log}&=\frac{1}{3B}\sum_{b,\theta}
 \log(1+m_{b\theta}/10^{-4}), \label{eq:image_log}\\
 \mathcal L_{\mathrm{rgb}}&=\operatorname{mean}
 \sqrt{(\widehat I-I)^2+10^{-6}}, \label{eq:image_rgb}\\
 \mathcal L_{\mathrm{ssim}}&=\frac{1}{3B}\sum_{b,\theta}
 [1-\operatorname{SSIM}(\widehat J_{b\theta},J_{b\theta})].
 \label{eq:image_ssim}
\end{align}
The stabilized logarithmic term balances errors across individual images and angles, the RGB Charbonnier loss preserves color fidelity, and the full-resolution SSIM~\cite{wang2004ssim} supervises local structure. All terms operate on continuous inverse-normalized predictions without clipping.

Given the observed $J_0$, we compute $\widehat S_1=J_0-\widehat J_{90}$ and $\widehat S_2=\widehat J_{45}-\widehat J_{135}$, with the reference quantities following Eqs.~\eqref{eq:stokes}--\eqref{eq:stokes_s2} in grayscale. The measured Stokes loss is then defined as
\begin{equation}
 \mathcal L_{\mathrm{Stokes}}=\tfrac12\left(
 \langle|\widehat S_1-S_1|\rangle+
 \langle|\widehat S_2-S_2|\rangle\right),
 \label{eq:stokes_loss}
\end{equation}
where $\langle\cdot\rangle$ denotes averaging over pixels and samples. The $S_1$ term reduces to grayscale L1 supervision of the $90^\circ$ view, whereas the $S_2$ term explicitly couples the $45^\circ$ and $135^\circ$ predictions. To further supervise relative polarization modulation, we define the normalized Stokes vector $v=(q,u)=(S_1,S_2)/\max(S_0,0.05)$ and its predicted counterpart $\widehat v$, with $\widehat S_0=J_0+\widehat J_{90}$. Valid pixels are those with measured $S_0\geq0.05$, no grayscale analyzer value at or above $250/255$, and measured $\rho\leq1$.

We divide the valid pixels into GT DoLP intervals $[0,0.05)$, $[0.05,0.1)$, $[0.1,0.2)$, and $[0.2,1]$. Let $\mathcal R_w$ denote the average of the ordinary valid-pixel mean and a weighted mean of these interval means. The vector weights $(0.10,0.15,0.25,0.50)$ increase the contribution of stronger modulation. For direction supervision, only pixels with $\rho\geq0.1$ are used, with weights $(0.25,0.75)$ for the last two intervals, thereby avoiding emphasis on unstable weak-polarization orientations. With $c(x)=\sqrt{x^2+10^{-6}}-10^{-3}$, the auxiliary losses are
\begin{align}
 \mathcal L_v&=\mathcal R_w\!\left[\tfrac12\{c(\widehat q-q)+c(\widehat u-u)\}\right], \label{eq:polar_vector}\\
 \mathcal L_a&=\mathcal R_w\!\left[1-\frac{\widehat v^{\mathsf T}v}
 {\sqrt{\|\widehat v\|^2+10^{-4}}\max(\rho,10^{-6})}\right], \label{eq:polar_direction}\\
 \mathcal L_m&=\left\langle c\!\left(\sqrt{\|\widehat v\|^2+10^{-8}}
 -\sqrt{\|v\|^2+10^{-8}}\right)\right\rangle_{\mathrm{valid}}.
 \label{eq:polar_magnitude}
\end{align}
The cosine ratio is clipped to $[-1,1]$. Interval weights are renormalized over bins containing at least 256 valid pixels per batch, while smaller bins remain in the ordinary mean. The complete objective is
\begin{equation}
 \mathcal L=\mathcal L_{\mathrm{img}}+0.1\mathcal L_{\mathrm{Stokes}}
 +\lambda_v\mathcal L_v+\lambda_a\mathcal L_a+\lambda_m\mathcal L_m,
 \label{eq:total_loss}
\end{equation}
with $(\lambda_v,\lambda_a,\lambda_m)=(0.0370,0.00370,0.00742)$. Together, these terms supervise polarization differences, normalized modulation, orientation, and magnitude.

\section{DATA PREPARATION AND IMPLEMENTATION DETAILS}
\label{sec:data}
In this section, we first provide a detailed introduction to our dataset in Section~\ref{subsec:4.1}, and then present the implementation details in Section~\ref{subsec:4.2}.

\subsection{Data Acquisition and Scene Coverage}
\label{subsec:4.1}
To support the development and evaluation of polarization image generation in real-world indoor and outdoor environments, we construct HighPol, a large-scale RGB-polarization dataset that exhibits rich polarization variations. Data were captured using a LUCID TRI050S1-QC polarization camera, whose micro-polarizer array samples four analyzer orientations in a single exposure. The camera was operated through ArenaView with a fixed exposure time of $47{,}178.6\,\mu\mathrm{s}$. Using the software's polarization-view mode, four-angle RGB images were exported at a resolution of $1224\times1024$ and subsequently resized to $512\times512$ for the generation experiments. The dataset covers diverse scenes and illumination conditions, including both natural and artificial lighting.

\begin{figure}[!t]
	\centering
	\includegraphics[width=\columnwidth]{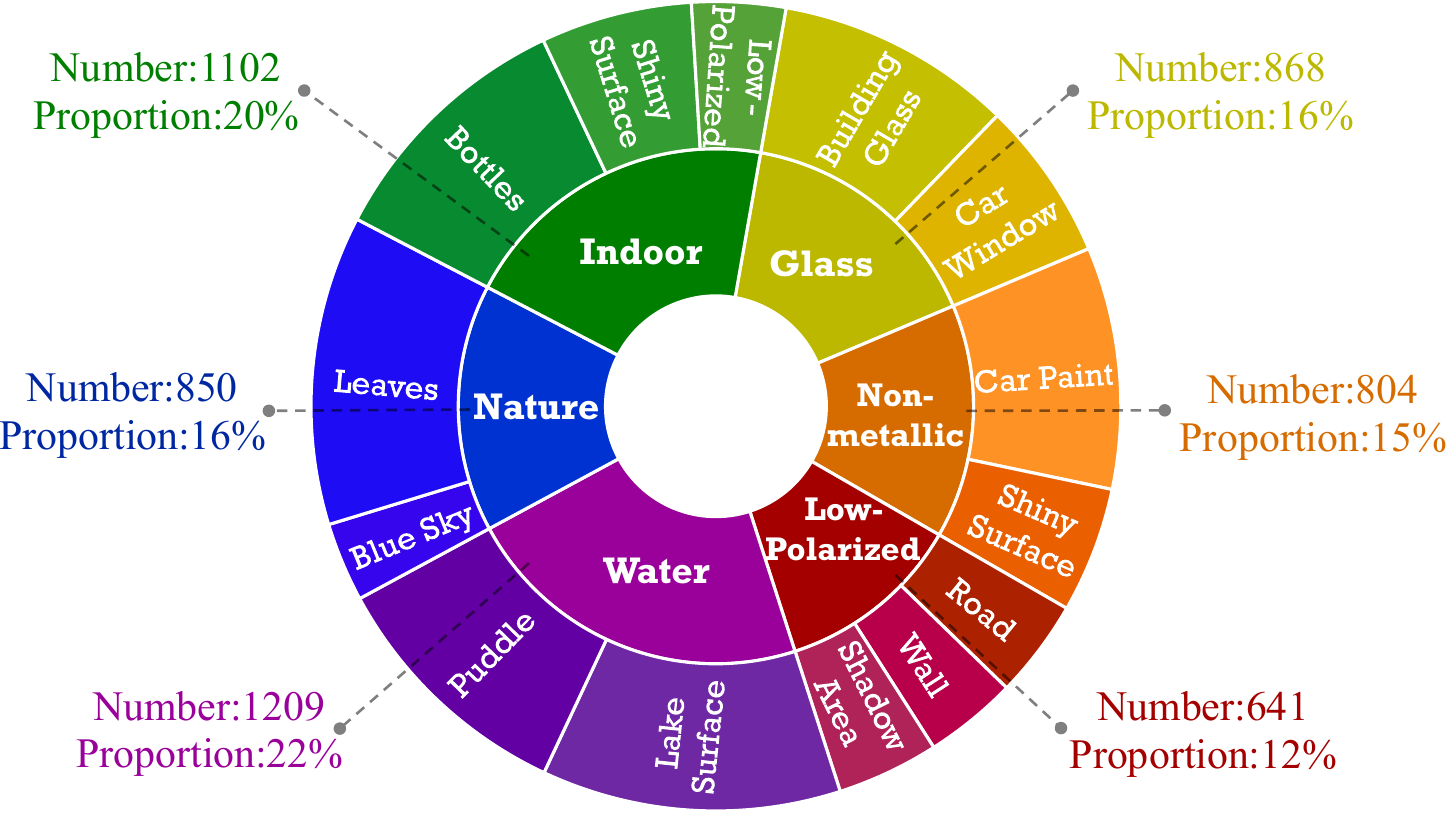}
	
	\vspace{2pt}
	{\footnotesize (a) Category distribution of the proposed HighPol dataset.\par}
	
	\vspace{4pt}
	\includegraphics[width=\columnwidth]{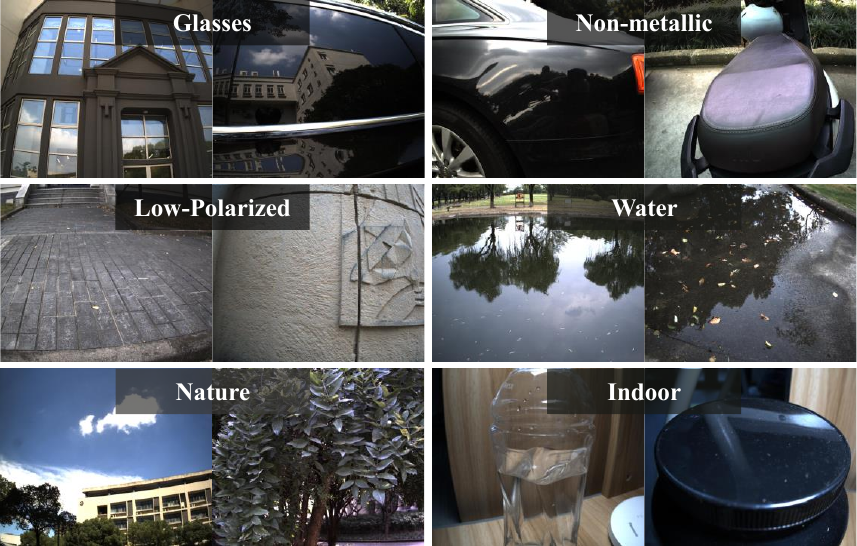}
	
	\vspace{2pt}
	{\footnotesize (b) Representative samples of six major categories.\par}
	
	\caption{Overview of the proposed HighPol dataset. (a) Category distribution of the proposed dataset. (b) Representative samples of six major categories, covering different materials, reflection characteristics, and polarization properties.}
	\label{fig:3}
\end{figure}

The HighPol dataset covers a broad range of indoor and outdoor scenes with diverse material reflection properties. As illustrated in Fig.~\ref{fig:3}(a), outdoor scenes include glass curtain walls, metallic vehicles, and water surfaces, whereas indoor scenes contain transparent plastics, ceramics, and tabletop materials. To enrich the distribution of polarization patterns, scenes were collected across different times of day (morning, noon, and evening) and weather conditions (sunny, cloudy, and rainy), leading to substantial variation in both polarization intensity and spectral appearance. In total, HighPol consists of 5,474 four-angle polarization image groups, corresponding to 21,896 high-resolution images. 

Table~\ref{tab:dataset_comparison} compares the proposed dataset with existing public polarization datasets. HighPol complements these resources with varied scene and material coverage at high spatial resolution. Specifically, scene diversity is measured by the variety of scene categories and polarization-sensitive materials, while polarization strength coverage is reflected by the richness of the DoLP distribution. These characteristics make the dataset suitable for both polarization image generation and polarization-aware perception. Fig.~\ref{fig:3}(b) shows representative examples with diverse polarization patterns and realistic reflection properties.

\begin{table}[!t]
	\caption{Comparison of existing polarization datasets. Diversity denotes scene-type coverage. DeepSfP size follows the authors' released-set description.}
	\label{tab:dataset_comparison}
	\centering
	\footnotesize
	\renewcommand{\arraystretch}{1.05}
	\setlength{\tabcolsep}{3pt}
	\begin{tabular*}{\columnwidth}{@{\extracolsep{\fill}}lccccc@{}}
		\hline
		\textbf{Dataset} & \textbf{Size} & \textbf{Resolution} & \textbf{Public} & \textbf{Diversity} & \textbf{Pol.strength} \\
		\hline
		PolaRGB~\cite{yao2025polarfree}      & 6500  & 1224$\times$1024 & Yes & Med.   & High \\
		DeepSfP~\cite{ba2020deep}      & 338   & 1224$\times$1024 & Yes & Low    & Med.  \\
		RGBP-Glass~\cite{mei2022glass}   & 4511 & 612$\times$512   & Yes & Low    & High \\
		Film Removal~\cite{tang2024learning} & 315   & 2448$\times$2048 & Yes & Low    & Med.  \\
		Kondo~\cite{kondo2020accurate}        & 44305 & 256$\times$192   & No  & Synth. & Low   \\
		\hline
		HighPol (Ours)              & 5474  & 1224$\times$1024 & Yes & High   & High \\
		\hline
	\end{tabular*}
\end{table}

\subsection{Implementation Details}
\label{subsec:4.2}

\noindent\textbf{Experimental settings.}
We combine 5,474 four-angle image groups from HighPol with 3,526 groups from PolarVis to construct a 9,000-group benchmark at $512\times512$ resolution. Samples are organized into clusters using contiguous source-index blocks and visual similarity computed from pretrained image features and image-level comparisons. Each cluster is assigned intact to a single partition, yielding 7,500 training, 750 validation, and 750 test samples.
MPIG is initialized from a public ImageNet-pretrained encoder and trained for 80 epochs using AdamW with a batch size of 14 and seed 42 on an A800 80\,GB GPU. The encoder and decoder learning rates are initialized to $10^{-5}$ and $10^{-4}$, respectively, with a five-epoch warmup followed by cosine decay. During epochs 1--4, the base image--Stokes objective is blended with an initialization objective using weights $e/5$ and $1-e/5$, respectively. The initialization objective combines residual and image reconstruction with Stokes-difference, intensity-sum residual, modulation-magnitude, and direction supervision. The vector, direction, and magnitude auxiliary losses use fixed coefficients from epoch 1 onward. From epoch 5, the initialization objective is removed and training follows Eq.~\eqref{eq:total_loss}.

\noindent\textbf{Comparison settings.}
All baselines use the same data split and receive $I_0$ as input. Pix2Pix~\cite{isola2017image} and the Swin generator~\cite{liu2021swin} are trained for 100 epochs. For the conditional DDPM~\cite{ho2020denoising}, we adopt a U-Net with lossless $2\times2$ pixel packing, 1,000 diffusion timesteps, velocity prediction, and 50-step DDIM sampling. It is trained for 120 epochs, and the epoch-100 EMA checkpoint is selected for evaluation. RGB2Pol-MAE~\cite{lin2025rgbpolarization} is trained for 150,000 updates with $224\times224$ crops and uses tiled inference. PolarAnything~\cite{zhang2025polaranything} is fine-tuned for 80,000 updates and evaluated using the selected 40,000-update checkpoint with 20 sampling steps. GenPolar~\cite{luo2026genpolar} is trained for 100 teacher epochs and 30 student-distillation epochs, with the selected teacher and student checkpoints at epochs 60 and 30, respectively; its student performs one-step inference. Baseline checkpoints are selected by validation image PSNR. PolarAnything and GenPolar are initialized from their public pretrained weights. To adapt the three polarization methods from RGB input to $I_0$, MAE and GenPolar predict RGB Stokes components, whereas PolarAnything predicts a polarization state shared across color channels. These predictions are then combined with $I_0$ to synthesize the missing views.

\noindent\textbf{Evaluation settings.}
For evaluation, predictions are clipped to $[0,1]$ and saved as 8-bit RGB images. Both predictions and references are then read as grayscale images using OpenCV and rescaled to $[0,1]$ for image and polarization assessment. Image PSNR and SSIM are computed for each predicted view and subsequently averaged over the three target angles and all 750 test groups. DoLP is computed with $S_0=J_0+J_{90}$ and a denominator stabilizer of $10^{-8}$. AoLP MAngE averages $\min(|\widehat{\phi}-\phi|,\pi-|\widehat{\phi}-\phi|)$ over all pixels and test groups. For either prediction or reference, pixels with $S_1=S_2=0$ are assigned the numerical value $\phi=0$ and remain included in the average. We report DoLP PSNR/SSIM/MAbsE and AoLP MAngE. The DoLP quality metrics use values clipped to $[0,1]$, whereas the DoLP$>1$ ratio is computed before this DoLP clipping. Both are derived from the same saved 8-bit analyzer images. Physical diagnostics include the mean intensity-sum residual $E_\Sigma=\langle|J_0+J_{90}-J_{45}-J_{135}|\rangle$ and the percentage of pixels with DoLP$>1$, denoted $r_{>1}$. The parameter count includes all instantiated networks, including unused pretrained modules in PolarAnything and GenPolar. FLOPs cover all three $512\times512$ outputs, including tiling and sampling, with two FLOPs per multiply--accumulate.

\section{EXPERIMENTS}
\label{sec:experiments}

In this section, we first compare MPIG with general image generators and task-adapted polarization estimators in Section~\ref{subsec:results}. We then evaluate its downstream utility in Section~\ref{subsec:apps}, followed by controlled ablations of the final architecture in Section~\ref{subsec:ablation}.

\subsection{Evaluation on HighPol/PolarVis}
\label{subsec:results}

\begin{figure*}[!t]
\centering
\includegraphics[width=\textwidth]{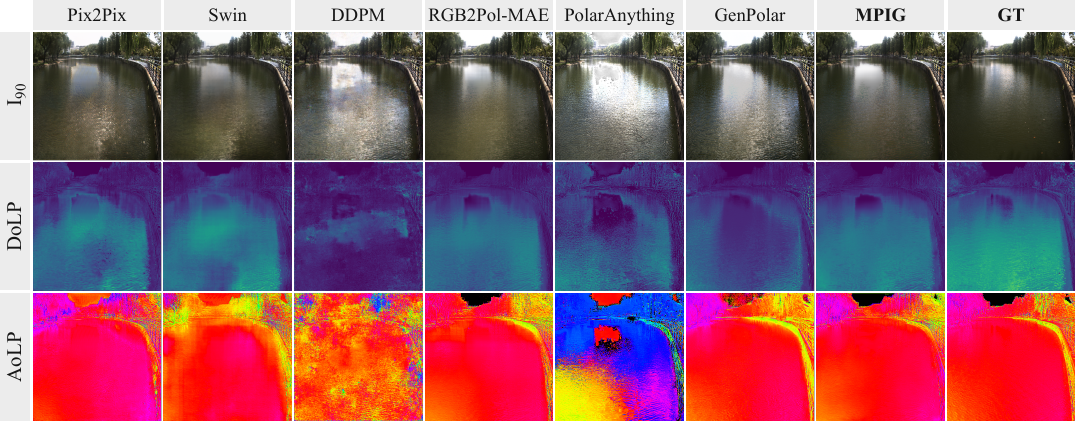}
\caption{Qualitative comparison on a HighPol water scene from the test set. Rows show the $90^\circ$ intensity image, DoLP, and AoLP; GT uses the measured four-angle images. DoLP and AoLP share display ranges of $[0,1]$ and $[0,180^\circ)$ across methods.}
\label{fig:results}
\end{figure*}

\begin{table*}[!t]
\centering
\caption{Comparison on the 750-group test set. Bold indicates the best image and polarization accuracy and the lowest computational cost among learned methods.}
\label{tab:comparison_methods}
\begingroup
\fontsize{9.5}{11.4}\selectfont
\setlength{\tabcolsep}{1.2pt}
\renewcommand{\arraystretch}{1.15}
\begin{tabular*}{\textwidth}{@{\extracolsep{\fill}}l*{10}{c}@{}}
\toprule
\multirow{2}{*}{Method} & \multicolumn{2}{c}{Image fidelity} & \multicolumn{3}{c}{DoLP} & AoLP & \multicolumn{2}{c}{Physical diagnostics} & \multicolumn{2}{c}{Complexity} \\
\cmidrule(lr){2-3}\cmidrule(lr){4-6}\cmidrule(lr){7-7}\cmidrule(lr){8-9}\cmidrule(l){10-11}
& PSNR$\uparrow$ & SSIM$\uparrow$ & PSNR$\uparrow$ & SSIM$\uparrow$ & MAbsE$\downarrow$ & MAngE$\downarrow$ & $E_\Sigma\downarrow$ & $r_{>1}\downarrow$ & Params. & FLOPs \\
& (dB) & & (dB) & & & ($^\circ$) & ($10^{-2}$) & (\%) & (M) & (T) \\
\midrule
Pix2Pix~\cite{isola2017image} & 27.715 & 0.9263 & 21.960 & 0.5525 & 0.06055 & 27.217 & 2.605 & 0.0863 & \textbf{34.26} & 0.498 \\
Swin~\cite{liu2021swin} & 26.994 & 0.9140 & 21.717 & 0.5298 & 0.06483 & 28.125 & 2.745 & 0.0788 & 39.21 & \textbf{0.287} \\
DDPM~\cite{ho2020denoising} & 24.952 & 0.8631 & 18.007 & 0.3521 & 0.09277 & 37.330 & 2.586 & 0.2570 & 63.37 & 18.940 \\
RGB2Pol-MAE~\cite{lin2025rgbpolarization} & 28.904 & 0.9437 & 23.174 & 0.6125 & 0.05129 & 21.718 & 0.153 & 0.0621 & 329.94 & 1.208 \\
PolarAnything~\cite{zhang2025polaranything} & 23.851 & 0.7912 & 18.332 & 0.1873 & 0.09150 & 33.947 & 0.255 & 0.0342 & 1788.79 & 46.003 \\
GenPolar~\cite{luo2026genpolar} & 27.717 & 0.9225 & 20.274 & 0.4835 & 0.06856 & 23.399 & 0.172 & 0.4367 & 1427.55 & 6.129 \\
\textbf{MPIG (ours)} & \textbf{29.729} & \textbf{0.9505} & \textbf{23.381} & \textbf{0.6413} & \textbf{0.05037} & \textbf{21.514} & 2.472 & 0.0345 & 85.86 & 1.126 \\
\midrule
Copy-$I_0$ & 26.508 & 0.9206 & 18.323 & 0.0852 & 0.11037 & --- & 0.000 & 0.0000 & 0 & 0 \\
Measured GT & --- & --- & --- & --- & --- & --- & 2.997 & 0.0474 & --- & --- \\
\bottomrule
\end{tabular*}
\endgroup
\end{table*}

\noindent\textbf{Image and polarization accuracy.}
As shown in Table~\ref{tab:comparison_methods}, MPIG achieves the best performance across all six image and polarization quality metrics. Relative to RGB2Pol-MAE, the strongest overall baseline, MPIG improves PSNR by 0.825\,dB and SSIM by 0.0068. It further increases DoLP PSNR from 23.174 to 23.381\,dB while reducing DoLP MAbsE and AoLP MAngE to 0.05037 and 21.514$^\circ$, respectively. In contrast, general image generators produce substantially larger angular errors, namely 27.217$^\circ$ for Pix2Pix, 28.125$^\circ$ for Swin, and 37.330$^\circ$ for DDPM. RGB2Pol-MAE and GenPolar lower these errors to 21.718$^\circ$ and 23.399$^\circ$, respectively, underscoring the benefit of explicitly modeling polarization quantities.

\noindent\textbf{Measurement-domain fidelity and physical diagnostics.}
The Copy-$I_0$ achieves 26.508\,dB PSNR by preserving the appearance shared across analyzer views. MPIG gains 3.220\,dB over this baseline and reduces DoLP MAbsE by 54.4\%, demonstrating its improved prediction of polarization modulation. By repeating the same observation, Copy-$I_0$ yields zero intensity-sum residual and zero DoLP. Its AoLP MAngE is omitted because its polarization orientation is undefined at every pixel. For the measured GT, $E_\Sigma=0.02997$ and $r_{>1}=0.0474\%$, whereas the corresponding MPIG values are 0.02472 and 0.0345\%. These diagnostics provide complementary evidence to the DoLP and AoLP prediction errors reported in Table~\ref{tab:comparison_methods}.

\noindent\textbf{Computational cost.}
MPIG uses 85.86\,M parameters and 1.126\,T FLOPs, reducing the parameter count and computation of RGB2Pol-MAE by 74.0\% and 6.8\%, respectively. The conditional DDPM requires 18.940\,T FLOPs for 50 sampling steps, while PolarAnything and GenPolar require 46.003\,T and 6.129\,T under their respective inference settings. Although Pix2Pix and Swin incur lower costs than MPIG, they also yield lower image and polarization accuracy. Consequently, MPIG combines the strongest quality scores in this comparison with lower computation than the three polarization baselines.

\noindent\textbf{Qualitative comparison.}
Fig.~\ref{fig:results} compares intensity and polarization predictions on a water scene. Pix2Pix and Swin tend to smooth the reflected structures, whereas DDPM and PolarAnything produce fragmented polarization patterns. GenPolar recovers a smoother spatial layout but underestimates the displayed DoLP over much of the water region. In contrast, MPIG preserves both the main reflection structure and the broad DoLP/AoLP patterns of the measured scene. Notably, the polarization maps reveal differences that are less apparent in the intensity images.

\subsection{Downstream Validation}
\label{subsec:apps}

\begin{figure*}[!t]
	\centering
	\includegraphics[width=\textwidth]{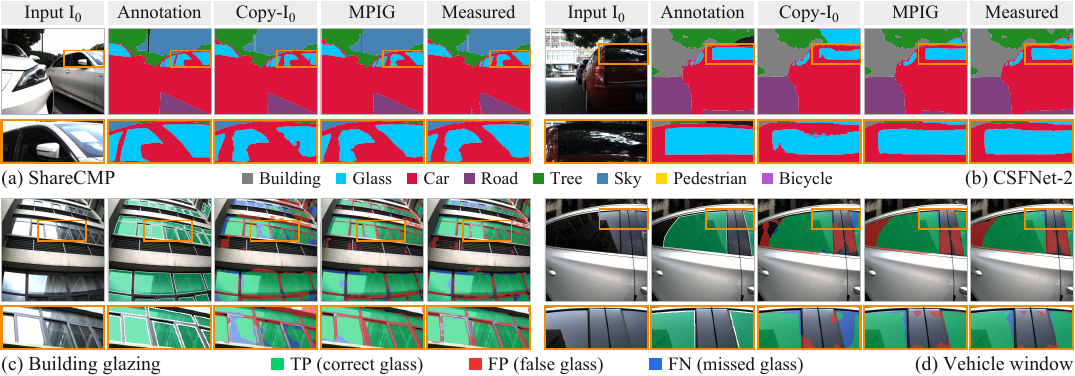}
	\caption{Downstream segmentation under frozen-input substitution. (a) ShareCMP and (b) CSFNet-2 on ZJU-RGB-P; (c) building glazing and (d) a vehicle window on HighPol. Within each panel, the intensity input and model weights are fixed. Annotation denotes the reference segmentation, whereas Measured denotes predictions using measured polarization inputs.}
	\label{fig:downstream}
\end{figure*}

\begin{table*}[t]
	\centering
	\caption{Downstream segmentation performance (\%). HighPol results are mean $\pm$ standard deviation across three seeds.}
	\label{tab:downstream_all}
	\begingroup
	\fontsize{9.5}{11.4}\selectfont
	\setlength{\tabcolsep}{1.5pt}
	\renewcommand{\arraystretch}{1.14}
	\begin{tabular*}{\textwidth}{@{\extracolsep{\fill}}l*{9}{c}@{}}
		\toprule
		& \multicolumn{6}{c}{ZJU-RGB-P} & \multicolumn{3}{c}{HighPol} \\
		\cmidrule(lr){2-7}\cmidrule(l){8-10}
		& \multicolumn{4}{c}{(a) Frozen-input substitution}
		& \multicolumn{2}{c}{(b) Input-specific training}
		& \multicolumn{3}{c}{(c) Frozen-input substitution} \\
		& \multicolumn{2}{c}{ShareCMP} & \multicolumn{2}{c}{CSFNet-2}
		& \multicolumn{2}{c}{ShareCMP} & \multicolumn{3}{c}{Glass model (three seeds)} \\
		\cmidrule(lr){2-3}\cmidrule(lr){4-5}\cmidrule(lr){6-7}\cmidrule(l){8-10}
		Input condition & mIoU & Glass IoU & mIoU & Glass IoU
		& mIoU & Glass IoU & Glass IoU & Dice & BF1 \\
		\midrule
		Measured & 91.967 & 87.158 & 91.405 & 86.114 & 91.967 & 87.158 & 68.09\,\(\pm\)\,1.09 & 80.13\,\(\pm\)\,0.94 & 22.83\,\(\pm\)\,1.30 \\
		MPIG & 91.536 & 85.889 & 90.445 & 83.811 & 92.042 & 87.074 & 65.56\,\(\pm\)\,4.27 & 78.26\,\(\pm\)\,3.32 & 21.76\,\(\pm\)\,1.67 \\
		Copy-$I_0$ & 91.392 & 84.998 & 89.162 & 79.538 & 91.866 & 86.343 & 51.80\,\(\pm\)\,18.89 & 65.88\,\(\pm\)\,18.01 & 14.72\,\(\pm\)\,4.35 \\
		\midrule
		MPIG $-$ Copy-$I_0$ & +0.144 & +0.891 & +1.283 & +4.272 & +0.176 & +0.731 & +13.75 & +12.38 & +7.05 \\
		\bottomrule
	\end{tabular*}
	\endgroup
\end{table*}

\noindent\textbf{Evaluation protocol.}
We evaluate the utility of MPIG-generated polarization views through controlled segmentation experiments. Specifically, three input conditions are considered: measured four-angle observations (Measured), measured $I_0$ combined with the MPIG-generated $45^\circ$, $90^\circ$, and $135^\circ$ views (MPIG), and four identical copies of $I_0$ (Copy-$I_0$). Across all conditions, the intensity branch consistently receives the same measured $I_0$. We distinguish between frozen-input substitution, which replaces the polarization inputs of a fixed downstream model, and input-specific training, which trains the same architecture separately under each input condition. Throughout the downstream experiments, MPIG is held fixed.

\noindent\textbf{Frozen-input substitution on ZJU-RGB-P.}
Following the official ZJU-RGB-P~\cite{xiang2021polarization} split of 344 training and 50 validation images, we evaluate ShareCMP~\cite{liu2025sharecmp} and CSFNet-2~\cite{qashqai2026csfnet}. ShareCMP employs a checkpoint selected on measured validation inputs, whereas CSFNet-2 uses its released checkpoint. Both checkpoints are evaluated under the three input conditions.
As shown in Table~\ref{tab:downstream_all}(a), MPIG improves mIoU over Copy-$I_0$ by 0.144 and 1.283 percentage points for ShareCMP and CSFNet-2, respectively. The corresponding Glass IoU gains are 0.891 and 4.272 points. As illustrated in Fig.~\ref{fig:downstream}(a,b), using the generated views leads to more complete segmentation of the vehicle windows. These improvements across both backbones indicate that MPIG recovers polarization cues beneficial for segmentation even with fixed downstream weights.

\noindent\textbf{Input-specific training with a shared architecture.}
To assess adaptation to generated inputs, we train three ShareCMP models with a common 200-epoch training recipe, using Measured, MPIG, and Copy-$I_0$ inputs, respectively. Each model contains 43.83 million parameters and is trained with seed 42, with checkpoints selected by validation mIoU under the corresponding input condition. As shown in Table~\ref{tab:downstream_all}(b), training and evaluating ShareCMP with MPIG-generated views achieves 92.042\% mIoU and 87.074\% Glass IoU, exceeding the separately trained Copy-$I_0$ model by 0.176 and 0.731 percentage points, respectively. Relative to feeding the generated views to the frozen measured-input model, input-specific training improves the two metrics by 0.506 and 1.185 points. The resulting performance is comparable to that of measured-input training, with a slightly higher mIoU and a Glass IoU difference of only 0.084 points. These results indicate that training directly on the generated views enables ShareCMP to better exploit the recovered polarization information.

\noindent\textbf{Glass segmentation on HighPol.}
We further evaluate a manually annotated subset containing 97 glass masks, which is divided into 40 training, 6 validation, and 51 test images. The splits are scene-disjoint, and all test images belong to the generator's held-out test split. A PGSNet-inspired network~\cite{mei2022glass} with intensity, AoLP, and DoLP branches is trained on measured inputs using seeds 42, 43, and 44. For each seed, the same model weights and validation-selected prediction threshold are applied across all three input conditions.
Table~\ref{tab:downstream_all}(c) reports the mean and standard deviation across seeds, computed from per-image metric averages. Boundary F1 is measured with a two-pixel tolerance at $512\times512$. Relative to Copy-$I_0$, MPIG improves Glass IoU, Dice, and Boundary F1 by 13.754, 12.377, and 7.046 percentage points, respectively. The scene-averaged Glass IoU gain remains positive across all six test scenes. As illustrated in Fig.~\ref{fig:downstream}(c,d), MPIG recovers glass regions that Copy-$I_0$ misses in both building glazing and vehicle windows. These results extend the gains observed on ZJU-RGB-P to glass segmentation in held-out HighPol scenes.

\subsection{Ablation Study}
\label{subsec:ablation}

\begin{figure}[!t]
\centering
\includegraphics[width=\columnwidth]{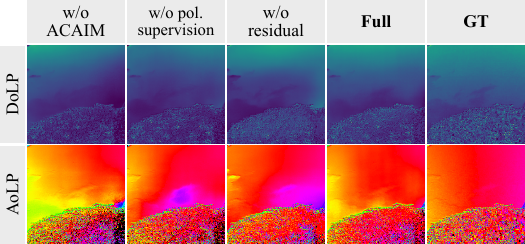}
\caption{Visual ablation on a reflective car window. Columns show MPIG without ACAIM, polarization supervision, or output residual prediction, followed by full MPIG and GT. DoLP (top) and AoLP (bottom) use a common $192\times192$ crop and the display ranges of Fig.~\ref{fig:results}.}
\label{fig:ablation}
\end{figure}

\begin{table}[!t]
\centering
\caption{Ablation of MPIG on the HighPol/PolarVis test set.}
\label{tab:ablation_mpig}
\begingroup
\fontsize{9.5}{11.4}\selectfont
\setlength{\tabcolsep}{1.5pt}
\renewcommand{\arraystretch}{1.15}
\begin{tabular*}{\columnwidth}{@{\extracolsep{\fill}}l*{4}{c}@{}}
\toprule
\multirow{2}{*}{Variant} & \multicolumn{2}{c}{Image} & DoLP & AoLP \\
\cmidrule(lr){2-3}\cmidrule(lr){4-4}\cmidrule(l){5-5}
& PSNR$\uparrow$ & SSIM$\uparrow$ & MAbsE$\downarrow$ & MAngE$\downarrow$ \\
\midrule
w/o ACAIM & 29.576 & 0.9499 & 0.05134 & 22.078 \\
w/o polar. supervision & 29.688 & 0.9502 & 0.05215 & 21.983 \\
w/o output residual & 29.469 & 0.9488 & 0.05175 & 23.161 \\
\textbf{Full MPIG} & \textbf{29.729} & \textbf{0.9505} & \textbf{0.05037} & \textbf{21.514} \\
\bottomrule
\end{tabular*}
\endgroup
\end{table}

Table~\ref{tab:ablation_mpig} evaluates the contributions of ACAIM, polarization-domain supervision, and output residual prediction. Removing ACAIM decreases image PSNR by 0.152\,dB and increases AoLP MAngE by 0.564$^\circ$, while DoLP MAbsE rises from 0.05037 to 0.05134. These results confirm the benefit of cross-angle feature exchange for jointly reconstructing the missing analyzer views. Fig.~\ref{fig:ablation} illustrates the corresponding changes in polarization patterns on a reflective surface.

When all explicit polarization-domain losses are removed, only image-domain supervision remains. This variant loses 0.041\,dB in image PSNR, while DoLP PSNR decreases by 0.229\,dB, DoLP MAbsE increases by 3.5\%, and AoLP MAngE rises by 0.469$^\circ$. Thus, polarization supervision improves the recovered modulation beyond the image reconstruction objective. We further replace GT DoLP grouping with generic difficult-pixel grouping based on image reconstruction error. The full model achieves 0.132\,dB higher DoLP PSNR and 0.123$^\circ$ lower AoLP MAngE, supporting polarization strength as a useful basis for allocating supervision.

Replacing output residual prediction with direct image prediction reduces PSNR by 0.260\,dB and raises AoLP MAngE by 1.647$^\circ$, which is the largest angular degradation among the three ablations. Residual prediction leverages the observed view to preserve shared scene appearance, allowing the decoders to focus on angle-dependent intensity changes. Its benefits extend to both image fidelity and polarization estimation, consistent with the DoLP and AoLP comparisons in Fig.~\ref{fig:ablation}.

We also compare the three angle-specific decoders with an alternative that uses two decoders to predict $S_1$ and $S_2$ and analytically synthesizes the missing views. With the same encoder, the Stokes-based alternative obtains 29.088\,dB PSNR and 22.721$^\circ$ AoLP MAngE, compared with 29.729\,dB and 21.514$^\circ$ for MPIG. This comparison favors direct residual decoding of the measured analyzer images over the tested analytical synthesis formulation.

\section{Conclusion}
\label{sec:conclusion}

In this work, we presented MPIG for predicting three missing analyzer views from a single $0^\circ$ observation through shared encoding, cross-angle residual decoding, and polarization-aware supervision. We further introduced HighPol, a large-scale dataset comprising 5,474 high-resolution four-angle image groups collected from diverse indoor and outdoor scenes. On the combined HighPol/PolarVis test set, MPIG achieves the best overall image and polarization quality among the evaluated methods, while requiring fewer parameters and FLOPs than RGB2Pol-MAE. Comparisons with the Copy-$I_0$ baseline and downstream segmentation experiments further demonstrate the utility of the predicted angle-dependent variations.

Despite these advances, polarization estimation remains challenging under strong reflections, severe intensity variations, and complex material responses. In addition, MPIG processes individual images and does not explicitly model temporal consistency. Future work will therefore focus on addressing these challenging conditions and extending the framework to temporally coherent and efficient video-based polarization reconstruction.

\bibliographystyle{IEEEtran}
\bibliography{references}
\end{document}